\documentclass[conference]{IEEEtran}
\usepackage{cite}
\usepackage{amsmath,amssymb,amsfonts}
\usepackage{algorithm}
\usepackage{graphicx}
\usepackage{textcomp}
\usepackage{xcolor}

\usepackage{url}
\usepackage{tikz}
\usetikzlibrary{positioning, shapes.geometric, arrows.meta}

\def\BibTeX{{\rm B\kern-.05em{\sc i\kern-.025em b}\kern-.08em
    T\kern-.1667em\lower.7ex\hbox{E}\kern-.125emX}}
\begin{document}

\title{Frugal Federated Learning for Violence Detection: A Comparison of LoRA-Tuned VLMs and Personalized CNNs}

\author{
	\IEEEauthorblockN{
    Sébastien Thuau\IEEEauthorrefmark{1}, 
    Siba Haidar\IEEEauthorrefmark{1}, 
    Ayush Bajracharya\IEEEauthorrefmark{2}, 
    Rachid Chelouah\IEEEauthorrefmark{2}}
\IEEEauthorblockA{\IEEEauthorrefmark{1}\textit{esieaLab}, 
        \textit{ESIEA}, 
        Paris, France \\
        Email: thuau@et.esiea.fr, siba.haidar@esiea.fr}
\IEEEauthorblockA{\IEEEauthorrefmark{2}\textit{ETIS Laboratory, CNRS, UMR8051}, 
	\textit{ University of CY Cergy}, 
	Paris, France \\
	Email: ayush.bajracharya@etu.cyu.fr, rc@cy-tech.fr}
}

\maketitle
\begin{abstract}
We examine frugal federated learning approaches to violence detection by comparing two complementary strategies: (i) zero-shot and federated fine-tuning of vision-language models (VLMs), and (ii) personalized training of a compact 3D convolutional neural network (CNN3D). Using LLaVA-7B and a 65.8M parameter CNN3D as representative cases, we evaluate accuracy, calibration, and energy usage under realistic non-IID settings.

Both approaches exceed 90\% accuracy. CNN3D slightly outperforms Low-Rank Adaptation(LoRA)-tuned VLMs in ROC AUC and log loss, while using less energy. VLMs remain favorable for contextual reasoning and multimodal inference. We quantify energy and CO$_2$ emissions across training and inference, and analyze sustainability trade-offs for deployment.

To our knowledge, this is the first comparative study of LoRA-tuned vision-language models and personalized CNNs for federated violence detection, with an emphasis on energy efficiency and environmental metrics.

These findings support a hybrid model: lightweight CNNs for routine classification, with selective VLM activation for complex or descriptive scenarios. The resulting framework offers a reproducible baseline for responsible, resource-aware AI in video surveillance, with extensions toward real-time, multimodal, and lifecycle-aware systems.

\end{abstract}


\section{Introduction}

Video surveillance systems increasingly rely on deep learning models for detecting and analyzing violent scenes in public spaces. However, centralized pipelines raise critical privacy concerns, especially when sensitive video data is collected and processed off-device. In parallel, the computational and environmental costs of deploying large models at scale have drawn scrutiny from both researchers and regulators.

This paper is part of the \textbf{DIVA} initiative, which stands for \textit{Decentralized Intelligence for Visual Awareness}. DIVA investigates sustainable, privacy-preserving AI methods for real-world video surveillance, with a focus on federated learning, multimodal reasoning, and energy-aware deployment.

Federated learning (FL) offers a promising solution by allowing models to be trained across multiple edge devices without sharing raw data, using secure aggregation and privacy-preserving protocols. Yet, FL alone does not resolve the challenge of model size and efficiency—particularly for multimodal architectures like vision-language models (VLMs), which may carry billions of parameters. We adopt “Vision-Language Models (VLMs)” as a general term for architectures that jointly process visual and textual inputs. This usage reflects the most common naming convention in the literature \cite{nguyen2024flora, alayrac2022flamingo, li2025benchmark}.

 In this study, we address two open questions: (1) whether large-scale VLMs can be frugally and effectively fine-tuned in federated non-IID settings, and (2) whether their deployment costs—especially in terms of energy and emissions—are justified when compared to lightweight CNNs that already perform competitively in violence classification tasks.

To this end, we conduct a systematic evaluation of centralized and federated learning strategies for video-based violence detection. We contrast zero-shot and Low-Rank Adaptation(LoRA)-adapted VLMs (e.g., LLaVA) with a compact 3D CNN trained under personalized federated learning. Our experiments quantify not only classification performance but also energy consumption and CO$_2$ emissions—providing practical insights for sustainable, privacy-aware deployment.

\section{Related Work}

% Federated Learning for Privacy-Aware Surveillance
\subsection{Privacy-Preserving Architectures for Video Surveillance}

Federated learning has been widely recognized as a key strategy to enhance privacy in video surveillance, where raw video data can contain highly sensitive personal information. Unlike centralized machine learning pipelines that require uploading raw footage to external servers, federated systems retain data locally and transmit only model updates or abstracted features. Several recent studies affirm this advantage.

Jiang et al.~\cite{jiang2022fedgnn} and Singh et al.~\cite{singh2022videoFL} describe architectures in which user or edge devices train local models, never exposing raw data, while secure aggregation protocols ensure privacy during synchronization. These are supported by large-scale surveys such as Abdulrahman et al.~\cite{abdulrahman2021survey} and Bellavista et al.~\cite{bellavista2021survey}, who discuss differential privacy, homomorphic encryption, ephemeral updates, and multi-party computation as layered privacy mechanisms.

Vyas and Torrens~\cite{vyas2024leaky} propose discarding raw video content post-processing and transmitting only AI-extracted motifs, which aligns with our work’s goal of frugal, privacy-aware deployment.

Overall, this body of work establishes federated learning as a privacy-enhancing alternative to centralized surveillance, with improved scalability, reduced network load, and multi-layered security—yet none explicitly address the cost-benefit trade-offs of federated fine-tuning for VLMs, which is the core contribution of this paper.

% Vision-Language Models and Zero-Shot Inference

% Conceptual Rationale for Federated and Frugal Approaches (NEW)

\subsection{LoRA Fine-Tuning in Federated Learning}
LoRA-based fine-tuning is theoretically well-suited to federated learning due to its parameter-efficient design, enabling significant reductions in communication and computation costs—critical factors in decentralized, resource-constrained environments.

\subsection{Energy-Efficient Multimodal AI Models}

Recent advances in personalized federated learning have introduced adapter-based fine-tuning as an efficient alternative to full model updates, particularly in low-resource or privacy-sensitive scenarios. LoRA (Low-Rank Adaptation) has emerged as a key mechanism in this context, enabling reduced communication and memory usage while maintaining competitive accuracy.

Below, we outline the key characteristics and empirical outcomes of recent LoRA-based federated learning frameworks:

\begin{itemize}
    \item \textbf{FLoCoRA}~\cite{ribeiro2024flocora}: Designed for vision models, it combines aggregation-agnostic LoRA adapters with affine quantization. This method achieves up to 63.9× reduction in communication with less than 4\% accuracy loss.

    \item \textbf{LoRA-FAIR}~\cite{bian2024lora-fair}: Integrates server-side correction and refined initialization strategies. It delivers performance improvements without incurring additional communication costs, outperforming prior state-of-the-art methods.

    \item \textbf{FLORA}~\cite{nguyen2024flora}: Applies rank-2 LoRA adapters to CLIP in federated settings. Reported gains include 30\% improvement in accuracy, 34.7× training speedup, and 2.47× reduction in memory usage across 16 vision datasets. \textbf{FLORA} represents the closest multimodal approach to our use case of federated surveillance, but it has not been evaluated on real-world video violence detection tasks.

    \item \textbf{FedLoRA}~\cite{yi2023fedlora}: Addresses client heterogeneity using a shared set of homogeneous LoRA adapters. Demonstrates 11.8× computational efficiency and a +1.35\% accuracy gain.

    \item \textbf{FDLoRA}~\cite{qi2024fdlora}: Employs a dual-adapter architecture that fuses global and personalized LoRA components. Achieves the highest mean accuracy across evaluated settings, emphasizing local adaptability without sacrificing global coherence.

    \item \textbf{HAFL}~\cite{su2024hafl}: Explores adaptive-rank tuning of LoRA adapters, which provides low communication overhead while avoiding performance degradation, making it suitable for highly constrained environments.
\end{itemize}

These contributions can be classified into two methodological families, Fig.~\ref{fig:taxonomy}:
\begin{itemize}
    \item \textbf{Aggregation-enhancing strategies}: FLORA and LoRA-FAIR improve server-side coordination, particularly during model aggregation.
    \item \textbf{Adapter-level innovations}: FDLoRA, FLoCoRA, HAFL, and FedLoRA emphasize efficient local tuning through quantization, personalization, and modular design.
\end{itemize}

% 5 taxonomy diagram of the methods

\tikzstyle{box} = [rectangle, draw, rounded corners, minimum height=1cm, minimum width=2.0cm, align=center, font=\footnotesize]
\tikzstyle{arrow} = [thick, ->, >=stealth]

\begin{figure}[htbp]
\centering
\resizebox{0.47\textwidth}{!}{%
\begin{tikzpicture}[node distance=1.3cm and 2.0cm]

% Root
\node (root) [box, fill=gray!20] {LoRA-Based Federated Learning};

% First layer
\node (agg) [box, below=.8cm of root, xshift=-2.0cm] {Aggregation\\Strategies};
\node (adpt) [box, below=.8cm of root, xshift=2.0cm] {Adapter\\Tuning};

% Aggregation strategies
\node (correction) [box, below=.8cm of agg, xshift=-1.2cm] {Server\\Correction\\(LoRA-FAIR)};
\node (stacking) [box, below=.8cm of agg, xshift=1.2cm] {Stacking\\(FLORA)};

% Adapter tuning strategies
\node (dual) [box, below=2.0cm of adpt, xshift=-2.4cm] {Dual\\Adapters\\(FDLoRA)};
\node (quant) [box, below=2.0cm of adpt] {Quantized\\LoRA\\(FLoCoRA)};
\node (adaptive) [box, below=2.0cm of adpt, xshift=2.4cm] {Adaptive\\Rank\\(HAFL)};

% Connections
\draw[arrow] (root) -- (agg);
\draw[arrow] (root) -- (adpt);

\draw[arrow] (agg) -- (correction);
\draw[arrow] (agg) -- (stacking);

\draw[arrow] (adpt) -- (dual);
\draw[arrow] (adpt) -- (quant);
\draw[arrow] (adpt) -- (adaptive);

\end{tikzpicture}
}
\caption{Taxonomy of LoRA-based federated learning approaches. Methods are grouped into (left) aggregation-side coordination (\emph{LoRA-FAIR}, \emph{FLORA}) and (right) client-side adapter tuning (\emph{FDLoRA}, \emph{FLoCoRA}, \emph{HAFL}). Our implementation follows the adapter-tuning path with standard LoRA modules on VLMs. The diagram is conceptual (no ranking implied).}

\label{fig:taxonomy}
\end{figure}
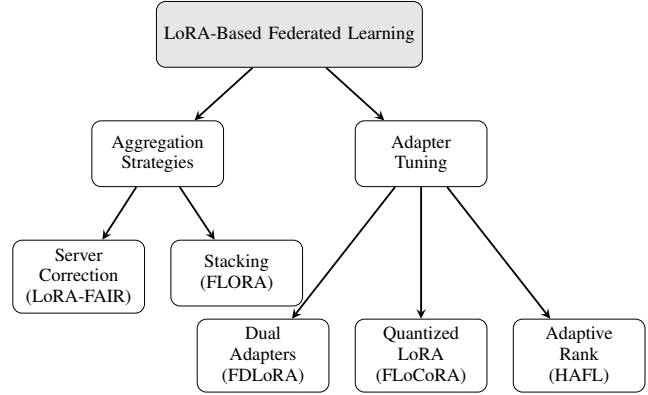

% 3. How to align with your contribution?
% Make this clear at the end of your SOTA section:

Despite promising results, none of the reviewed studies evaluate LoRA-based federated fine-tuning of VLMs for violence detection in realistic video surveillance settings, nor do they compare their cost-effectiveness against lightweight CNNs under identical non-IID scenarios. While several frameworks demonstrate strong potential for efficiency and personalization, their application to federated multimodal fine-tuning in video surveillance remains largely untested. In contrast to prior work focused on general vision or multimodal tasks, our study fills this gap by applying federated LoRA fine-tuning to VLMs in video-based violence detection—benchmarking both performance and environmental cost against a lightweight CNN baseline under non-IID conditions.

Beyond algorithmic considerations, the sustainability of AI models—particularly their energy and emissions footprint—is increasingly central to deployment decisions.

\subsection{Environmental Impact Assessment}

The environmental impact of AI has become a major societal concern, particularly as deep learning models are increasingly deployed in large-scale systems such as violence detection. A critical issue is the lack of transparency regarding the energy consumption and carbon emissions of various model architectures. This gap in knowledge hinders efforts to optimize AI systems for sustainability. Initiatives like the \textit{EU AI Act} and \textit{AI Summit} emphasize the need for robust regulatory frameworks to address these challenges, ensuring that AI technologies are developed and deployed with consideration for their environmental footprint. In France, projects like \textit{The Shift Project}, supported by the \textit{ADEME} (French Environment and Energy Management Agency), and \textit{AFNOR SPEC} standards (e.g. \textit{AFNOR SPEC 2314}), are actively working on defining best practices and guidelines for reducing the carbon impact of AI, promoting energy-efficient AI model development and fostering sustainable innovation in the field.

%Ajout connaissance de l'espoir recherche

This paper responds to these emerging regulatory and scientific demands by quantifying the energy use and emissions of centralized and federated fine-tuning strategies for VLMs and benchmarking them against frugal CNN baselines, thus informing future deployment choices in sustainable surveillance.
\section{Methodology}

\subsection{Overview}

Our study evaluates the trade-offs between personalized federated learning of lightweight CNN models~\cite{kassir2025, pajon2024cnn3d} and LoRA-based fine-tuning of vision-language models (VLMs) in decentralized video surveillance. We perform three experiments on a merged dataset (RWF-2000 + RLVS) under various training and inference settings, measuring performance and energy efficiency. We simulate realistic federated non-IID scenarios with controlled heterogeneity in video content and label distribution.

\subsection{Datasets and Client Partitioning}

We construct a combined dataset of 4000 labeled videos by merging the RWF-2000~\cite{rwf2000} and RLVS~\cite{rlvs2020} datasets, each contributing 2000 videos with binary labels (violent vs. non-violent). An 80/20 stratified split is used for training and validation.

To simulate federated non-IID conditions and contextual heterogeneity, we partition the data across 10 clients: the first five clients receive only RWF-2000 videos, and the remaining five only RLVS. This structure reflects real-world variation in surveillance sources (e.g., shopping mall vs. subway) and enforces label and domain skew. For one experiment, we additionally use a Dirichlet distribution with concentration parameter $\alpha=1$ to vary the degree of label imbalance within clients.

\subsection{Federated Learning Setup}

All federated training uses the FedAvg algorithm~\cite{mcmahan2017communication}, with 20 communication rounds. Experiments are conducted using a server-side simulation of federated learning: instead of deploying across physical edge devices, we emulate 10 independent clients on a single GPU server. Each client maintains isolated data partitions and local update procedures, and communication dynamics are simulated to reflect realistic federated behavior.

We simulate two levels of partial participation:
\begin{itemize}
    \item \textbf{50\% client fraction:} 5 out of 10 clients randomly selected per round.
    \item \textbf{30\% client fraction:} 3 out of 10 clients randomly selected per round.
\end{itemize}
Partial participation is used to mirror bandwidth and availability constraints common in real-world FL deployments, while reducing cumulative training cost.

\subsection{LoRA-Based Fine-Tuning: Centralized vs. Federated}

We apply LoRA~\cite{hu2021lora} fine-tuning to the \texttt{llava-hf/LLaVA-NeXT-Video-7B-hf}, a multimodal vision-language model using CLIP as the visual encoder. Videos are converted into representative frames (default: 24 frames per clip), which are passed through CLIP’s visual backbone. Prompts such as “\texttt{Analyze the video. Is this a fight scene? Answer yes or no.}” are used for both training and zero-shot evaluation.

Fine-tuning is conducted under two settings:
\begin{itemize}
    \item \textbf{Centralized LoRA:} LLaVA is fine-tuned on the full training set using a standard supervised objective.
    \item \textbf{Federated LoRA:} The same model is fine-tuned using federated LoRA with 4-bit quantization. Each client updates local LoRA adapters while keeping base model weights frozen. Communication involves only adapter weights.
\end{itemize}

LoRA adapters are initialized with rank 8 and scaling factor $\alpha=32$ using the Hugging Face \texttt{peft} library. The adapter layers are inserted into attention blocks of the LLaVA architecture.

\subsection{Personalized CNN3D with Parameter Decoupling}

For the baseline, we implement a custom 3D CNN model with approximately 66 million parameters. Instead of standard FedAvg, we apply personalized federated learning (PFL) using parameter decoupling. Specifically, all layers after spatiotemporal feature extraction are decoupled and trained locally per client, while shared layers are aggregated globally. This balances generalization with client-specific specialization, which is particularly valuable under domain heterogeneity (RWF vs. RLVS).

\subsection{Evaluation Protocol: Performance and Environmental Metrics}

We assess each method across three complementary experiments:
\begin{enumerate}
    \item \textbf{Exp. 1: Zero-shot inference with VLMs.} Pretrained models such as LLaVA, Qwen-VL, and Flamingo are evaluated directly without fine-tuning.
    \item \textbf{Exp. 2: Federated fine-tuning of LLaVA with LoRA.}
    \item \textbf{Exp. 3: Personalized federated training of CNN3D.}
\end{enumerate}

For each experiment, we collect:
\begin{itemize}
    \item \textbf{Performance metrics:} accuracy, F1 score, and video classification throughput.
    \item \textbf{Communication metrics:} number of rounds, model upload/download size, and adapter payload.
    \item \textbf{Environmental metrics:} energy consumption (kWh) and carbon emissions (gCO$_2$e), estimated using the \texttt{CodeCarbon} library~\cite{codecarbon}. Measurements are averaged across three independent runs. CO$_2$ estimates use a grid intensity of 42 gCO$_2$e/kWh (Paris region).
\end{itemize}

All training and inference are conducted on NVIDIA A10 GPUs. Default values are used for batch size and frame resolution. Early stopping is applied using a patience of 5 rounds based on validation F1 score.

\vspace{1mm}
\textit{On Energy Modeling.} While this study relies on direct energy measurements, we acknowledge the need for a formal modeling approach that can generalize beyond specific hardware or regional conditions. To this end, we adopt a conceptual framework in which the environmental impact of training or inference consists of two components: (i) operational emissions derived from electricity consumption, grid carbon intensity, and data center efficiency (e.g., Power Usage Effectiveness), and (ii) amortized embodied emissions associated with hardware manufacturing, proportionally allocated based on usage and lifetime.

In this framework, electricity consumption can be approximated using hardware-level specifications such as Thermal Design Power (TDP), utilization rate, duration of workload, and an overhead multiplier to account for system components outside the GPU. This formulation—presented during GT PIA internal working group meetings and inspired by prior sustainable AI methodology, with a formal publication pending—is consistent with models proposed in the sustainable AI literature.

However, for large vision-language models such as LLaVA NeXT Video and Ovis2, no published studies currently provide complete and validated measurements of utilization profiles, workload durations, or embodied emissions. While some isolated benchmarks report inference-level energy draw, the lack of open and detailed lifecycle data precludes full modeling at this stage.

Accordingly, our reported emissions are derived from empirical measurement only. We introduce the above modeling structure as a scientific foundation for future extensions, particularly in scenarios where direct measurement is not feasible or when projecting across heterogeneous deployment infrastructures.

\section{Experiments and Results}\label{sec:experiments}

\subsection{Exp. 1 : Zero-Shot VLM Inference: Accuracy vs. Efficiency}

In this experiment, we evaluate the performance and energy efficiency of large vision-language models (VLMs) used in a zero-shot setting—i.e., without any fine-tuning—on surveillance video classification. This scenario simulates a real-world deployment where pretrained models are directly queried to detect violence using fixed prompts such as: ``\texttt{Analyze the video. Is this a fight scene? Answer yes or no.}''

\subsubsection{Evaluation on RLVS.}
Table~\ref{tab:zeroshot-rlvs} presents the results on the full RLVS dataset (2000 videos). Ovis2-8B yields the highest accuracy (95.80\%) and F1 score (96\%) while maintaining a relatively low energy consumption of 457 Wh. Emissions can be derived from energy using 42 g CO$_2$/kWh. LLaVA-7B follows closely in accuracy (94.15\%) with slightly lower energy usage (328 Wh), highlighting a strong trade-off between precision and frugality. Other models such as Qwen-VL-7B and InternVL3 perform well in accuracy but lack complete environmental metrics. These results demonstrate that off-the-shelf VLMs can achieve high-quality violence detection on RLVS without model adaptation, though model selection remains critical for energy-sensitive deployments.

% table 1 Zero-Shot Inference on RLVS (2000 videos)
\begin{table}[ht]
\caption{Measured zero-shot inference on RLVS (2{,}000 videos); mean~$\pm$~std over 3 runs}
\centering
\scriptsize
\begin{tabular}{|l|c|c|c|}
\hline
\textbf{Model} & \textbf{Acc. (\%)} & \textbf{F1} & \textbf{Energy (Wh)}  \\
\hline
\textbf{Ovis2-8B}   & \textbf{95.80 ± 0.15} &\textbf{0.960 ± 0.004} & \textbf{457 ± 7}\\
LLaVA-7B            & 94.15 ± 0.20         & 94.0 ± 5          & 328 ± 6\\
Qwen-VL-7B          & 91.00 ± 0.25         & 91.0 ± 6          & 616 ± 8\\
InternVL3           & 92.75 ± 0.22         & 92.0 ± 5          & –                    \\
MiniCPM             & 87.75 ± 0.30         & 86.0 ± 6          & –                    \\
Ovis2-4B            & 60.75 ± 0.45         & –                & –                       \\
\hline
\multicolumn{4}{l}{\parbox[t]{0.9\linewidth}{– indicates missing or unmeasured values. \\Energy measured in Watt-hours (Wh). \\Emissions can be estimated using 42 g CO$_2$/kWh = 0.042 g/Wh.}
}
\end{tabular}
\vspace{1mm}
\label{tab:zeroshot-rlvs}
\end{table}

\subsubsection{Evaluation on RWF-2000 (interpolated).}
Since full inference runs on RWF-2000 were not initially available, we extrapolated the results from 400-video benchmarks using linear scaling, which is reasonable given the uniform number of frames per video and deterministic processing pipeline. Table~\ref{tab:zeroshot-rwf} shows the estimated results for Qwen2.5-Omni, Qwen-VL, and LLaVA-7B. Notably, LLaVA-7B retains strong accuracy (86.75\%) with the lowest energy usage (474 Wh) among the tested models. However, performance is weaker compared to RLVS—highlighting that zero-shot VLMs are sensitive to domain shift and dataset-specific content, which is especially relevant in surveillance contexts.

% table 2 Zero-Shot Inference on RWF-2000 (Interpolated)
\begin{table}[ht]
\caption{Extrapolated zero-shot inference on RWF-2000, based on 400-video test set, mean ± std over 3 runs}
\centering
\scriptsize
\begin{tabular}{|l|c|c|c|c|}
\hline
\textbf{Model} & \textbf{Acc. (\%)} & \textbf{F1 Score (\%)} & \textbf{Energy (Wh)} \\
\hline
Qwen2.5-Omni-7B    & 86.75 ± 0.25 & 86;75 ± 6 & 2013 ± 30 \\
Qwen2.5-VL-7B      & 85.75 ± 0.30 & 85.75 ± 6 & 872 ± 20 \\
\textbf{LLaVA-7B}  & \textbf{86.75 ± 0.20} & \textbf{86.75 ± 5} & \textbf{474 ± 10} \\
\hline
\multicolumn{4}{l}{\parbox[t]{0.9\linewidth}{Values scaled from 400-video benchmark using linear extrapolation. \\Energy measured in Watt-hours (Wh). \\Emissions can be estimated using 42 g CO$_2$/kWh = 0.042 g/Wh.}}
\end{tabular}
\vspace{1mm}

\label{tab:zeroshot-rwf}
\end{table}

\subsubsection{Combined benchmark on RLVS + RWF (800 videos).}
To enable a broader comparison of zero-shot inference performance under consistent conditions, we evaluated four vision-language models—Ovis2-8B, LLaVA-7B, InternVL3, and Qwen-VL—on a shared benchmark of 800 surveillance videos. As shown in Table~\ref{tab:excel-zeroshot}, Ovis2-8B achieves the highest accuracy (92.5\%), F1 score (92.48\%), and ROC AUC (92.5\%), while maintaining relatively low inference time and energy usage (1928 s, 172 Wh).

LLaVA-7B performs competitively, especially in terms of F1 (90.62\%), but with higher inference time and energy (3215 s, 200 Wh). InternVL3 and Qwen-VL also demonstrate good ROC AUC scores (around 0.885), but with slightly lower accuracy and, in the case of Qwen-VL, a substantially higher energy footprint (660 Wh) and runtime (5770 s). These results illustrate the trade-offs between predictive performance and efficiency across VLMs, and motivate the selection of LLaVA-7B for further fine-tuning experiments due to its strong balance of performance and adaptability.

% new table 3 Comparison zero shot Ovis LLaVA on 400 rwf +400 rlvs
\begin{table}[ht]
\caption{Zero-shot inference: measured performance on 400 videos from RLVS + 400 videos from RWF-2000 (mean ± std over 3 runs)}
\centering
\scriptsize
\resizebox{\linewidth}{!}{%
\begin{tabular}{|l|c|c|c|c|c|}
\hline
\textbf{Model} & 
\begin{tabular}{@{}c@{}}\textbf{Accuracy} \\ (\%)\end{tabular} & 
\begin{tabular}{@{}c@{}}\textbf{F1 Score} \\ (\%)\end{tabular} & 
\begin{tabular}{@{}c@{}}\textbf{ROC AUC} \\ (\%)\end{tabular} & 
\begin{tabular}{@{}c@{}}\textbf{Energy} \\ (Wh)\end{tabular} & 
\begin{tabular}{@{}c@{}}\textbf{Duration} \\ (s)\end{tabular} \\
\hline
Ovis2-8B     & 92.5 ± 0.2  & 92.5 ± 0.4  & 92.5 ± 0.6  & 172 ± 4  & 1928 ± 30 \\
LLaVA-7B     & 90.4 ± 0.3  & 90.6 ± 0.5  & 86.4 ± 0.7  & 200 ± 5  & 3215 ± 46 \\
InternVL3    & 88.5 ± 0.3  & 88.3 ± 0.4  & 88.5 ± 0.6  & 130 ± 4  & 1745 ± 35 \\
Qwen-VL      & 88.5 ± 0.3  & 87.4 ± 0.5  & 88.5 ± 0.5  & 660 ± 12 & 5770 ± 55 \\
\hline
\end{tabular}
}
\label{tab:excel-zeroshot}
\end{table}

\subsubsection{Takeaways and Summary}
Zero-shot VLMs such as Ovis2-8B, LLaVA-7B, InternVL3, and Qwen-VL demonstrate strong performance without fine-tuning. However, accuracy gaps and energy variability highlight the challenge of domain sensitivity—especially in non-IID surveillance environments. To address this, we explore federated fine-tuning and client-side personalization in subsequent experiments.

While Ovis2-8B yields the highest zero-shot accuracy and calibration, we focus our fine-tuning experiments on LLaVA-7B due to its support for LoRA-based adaptation and its mature open-source ecosystem. This makes it a reproducible and scalable candidate for frugal adaptation strategies in federated contexts.

\subsection{Exp. 2: Federated Fine-Tuning of LLaVA with LoRA}

In this experiment, we evaluate the benefits of LoRA-based fine-tuning of the LLaVA-7B model under a federated learning setup. We compare its performance against the zero-shot baseline on a heterogeneous, non-IID client distribution, simulating real-world deployment conditions where surveillance content differs across locations.

Although all clients are simulated on a single server, the setup captures key aspects of edge-oriented environments—such as non-IID data, limited participation, and communication-efficient updates via LoRA. This controlled simulation provides a reproducible testbed for evaluating feasibility and trade-offs. Future work will explore deployment on actual edge hardware to validate these findings under real-world constraints.

The full dataset (RWF-2000 + RLVS = 4000 videos) is partitioned across 10 clients: clients 1–5 receive only RWF videos, and clients 6–10 receive only RLVS. A Dirichlet distribution ($\alpha$ = 1) introduces label imbalance. Each client uses 24-frame clips, and the data is split 80/20 for training and validation (800 videos in total). We simulate partial participation: 5 clients are selected per round (50\%) over 20 rounds, using FedAvg. Only LoRA adapter weights are updated and exchanged; the base LLaVA model remains frozen. CodeCarbon estimates are computed under the Île-de-France grid (PUE = 1).

Table~\ref{tab:llava-lora-finetune} summarizes the results. Compared to zero-shot inference, federated LoRA fine-tuning significantly improves calibration (log loss reduced from 0.707 to 0.535) and ROC AUC (85.93 → 91.24), while also yielding slight gains in accuracy (+0.25) and decision confidence. Energy consumption for fine-tuning remains reasonable at 570 Wh, compared to 200 Wh for zero-shot inference.

We also tested a stricter setting with only 3 clients per round (30\% participation). While this configuration improved ROC AUC further (to 92.59), it yielded slightly lower accuracy and F1 score. We therefore retain the 5-client setup as our main reference.

These results confirm that LoRA-based federated fine-tuning enhances adaptability and confidence under realistic non-IID settings, without requiring full model updates or high energy budgets.

\begin{table}[ht]
\caption{LLaVA fine-tuned with Federated Learning(FL) and LoRA (FedAvg, 5 clients/round) vs. Zero-Shot on 800-video validation set (RWF + RLVS)}
\centering
\scriptsize
\begin{tabular}{|p{3.3cm}|c|c|}
\hline
\textbf{Metric} & \textbf{LoRA FL} & \textbf{Zero-Shot} \\
\hline
Training Duration (s)     & 4741     & –        \\
Energy (Wh)               & 570      & 200      \\
Emissions (g CO$_2$e)     & 32       & 8.4      \\
Accuracy (\%)             & 90.87    & 90.62    \\
F1 Score (\%)             & 90.93    & 90.84    \\
ROC AUC  (\%)             & 91.24    & 85.93    \\
Log Loss                  & 0.535   & 0.706   \\
\hline
\multicolumn{3}{l}{\parbox[t]{0.9\linewidth}{Validation set = 800 videos (20\% of RWF + RLVS). \\Energy measured in Wh. Mean over 3 runs.}
}
\end{tabular}
\vspace{1mm}
% \small\textit{Validation set = 800 videos (20\% of RWF + RLVS). Energy measured in Wh. Mean over 3 runs.}
\label{tab:llava-lora-finetune}
\end{table}

\subsection{Exp. 3: Personalized Federated Learning with Lightweight CNN3D}

In this experiment, we evaluate a personalized federated learning (PFL) approach based on a compact 3D convolutional neural network (CNN3D) with approximately 65.8M parameters. Unlike LLaVA, which is fine-tuned using LoRA on a frozen backbone, CNN3D uses full end-to-end training but with a decoupled architecture: shared feature extraction layers are aggregated via FedAvg, while client-specific classification heads are trained locally.

The federated setup mirrors Exp. 2: 10 clients (5 with RWF, 5 with RLVS), 20 communication rounds, 5 clients per round. The dataset is split 80/20, and the validation set consists of 800 videos, drawn from the full non-IID distribution. CodeCarbon measurements are computed with consistent assumptions (Île-de-France region, PUE = 1), and training is performed on a single NVIDIA A10 GPU.

Table~\ref{tab:cnn3d-vs-llava-full} reports the results alongside the LLaVA zero-shot and LoRA-adapted models. CNN3D with PFL achieves 90.75\% accuracy and 0.9066 F1-score, closely matching LLaVA + LoRA. It also reaches the highest ROC AUC (92.59) and the lowest log loss (.546), indicating strong calibration and generalization under domain shift. Most notably, CNN3D consumes only 240 Wh during training—less than half the energy of LoRA fine-tuning—and emits just 10.1 g CO$_2$e.

These results confirm that compact CNN architectures with personalized updates can match the performance of transformer-based models, while consuming significantly fewer resources—even when accounting for the inference-only cost of zero-shot VLMs like LLaVA (200 Wh). All energy values in this comparison include full training (for CNN3D and LoRA) or full inference (for zero-shot), including model loading where applicable.

\begin{table}[ht]
\caption{Federated performance: CNN3D vs. LLaVA Zero-Shot vs. LLaVA + LoRA (20 rounds, 5 clients/round)}
\centering
\scriptsize
\setlength{\tabcolsep}{3pt}
\begin{tabular}{|l|c|c|c|}
\hline
\textbf{Metric} & \textbf{CNN3D} & \textbf{LLaVA ZS} & \textbf{LLaVA LoRA} \\
\hline
Train Time (s) & 3795 & – & 4741 \\
Energy (Wh) & 240 & 200 & 570 \\
CO$_2$ (g) & 10.1 & 8.4 & 24.0 \\
Acc. (\%) & 90.75 & 90.62 & 90.87 \\
F1 (\%) & 90.66 & 90.84 & 90.93 \\
AUC & \textbf{92.59} & 85.93 & 91.24 \\
LogLoss & \textbf{5.46} & 7.07 & 5.35 \\
\hline
\multicolumn{4}{l}{\parbox[t]{.9\linewidth}{ZS = zero-shot. LoRA = LoRA fine-tuned. Acc. = Accuracy, AUC = ROC AUC. CNN3D and LLaVA LoRA are evaluated on a non-IID validation set (800 videos); LLaVA ZS is evaluated on a balanced, centralized 800-video test set.}}
\end{tabular}
\vspace{1mm}
\label{tab:cnn3d-vs-llava-full}
\end{table}

\subsection{Energy and Carbon Footprint Analysis}

We now analyze the energy consumption and CO$_2$ emissions across the three evaluated strategies: zero-shot inference with vision-language models (VLMs), federated LoRA fine-tuning of LLaVA, and personalized federated training of a lightweight CNN3D model.

From Table~\ref{tab:zeroshot-rlvs}, LLaVA-7B demonstrates efficient zero-shot inference with 200 Wh consumption and 8.4 g CO$_2$ emissions. By contrast, Qwen2.5-Omni-7B exhibits much higher energy usage (201 Wh) and emissions (113 g). Notably, Ovis2-8B achieves the best accuracy while remaining frugal (457 Wh), confirming that VLM energy profiles vary significantly—underscoring the importance of model selection for sustainable deployment.

In federated training (Table~\ref{tab:llava-lora-finetune}), LLaVA fine-tuned with LoRA consumes 570 Wh and emits 24 g CO$_2$e over 20 rounds across 10 clients. This cost is justified by improved calibration (log loss) and ROC AUC (85.93 → 91.24), enabling more confident and reliable predictions in real-world, heterogeneous settings.

By comparison, CNN3D with personalized updates (Table~\ref{tab:cnn3d-vs-llava-full}) delivers nearly equivalent accuracy while consuming only 240 Wh and emitting 10.1 g CO$_2$e—less than half of LoRA’s footprint. It also achieves the best ROC AUC (92.59) and lowest log loss (.546), suggesting that lightweight CNNs with decoupled updates offer both energy efficiency and strong generalization.

We summarize this trade-off in Table~\ref{tab:energy-summary}.

\begin{table}[ht]
\caption{Summary of energy and emissions across evaluated strategies}
\centering
\scriptsize
\begin{tabular}{|p{2.9cm}|c|c|c|}
\hline
\textbf{Method} & \textbf{Accuracy (\%)} & \textbf{Energy (Wh)} & \textbf{CO$_2$ (g)} \\
\hline
Zero-shot (LLaVA-7B)       & 90.62 & 200 & 8.4 \\
Fed LoRA (LLaVA-7B)        & \textbf{90.87} & 570 & 24.0 \\
PFL CNN3D (65.8M params)   & 90.75 & \textbf{240} & \textbf{10.1} \\
\hline
\end{tabular}
\label{tab:energy-summary}
\end{table}

These insights underscore the strategic value of model selection in resource-constrained federated environments. For routine classification tasks, personalized CNNs deliver strong accuracy with minimal energy and emissions. For nuanced inference or multimodal interpretability, VLMs with LoRA remain a viable—though more resource-intensive—choice. This duality supports a hybrid deployment model: use lightweight CNNs by default, and reserve VLMs for selective, high-context queries.

\subsection{Reflections on Lifecycle Energy Attribution and Transfer Overhead}

While our evaluation focuses primarily on inference and federated fine-tuning energy consumption, a complete environmental assessment would ideally account for the upstream cost of pretraining large vision-language models (VLMs). Training foundational models such as LLaVA-7B typically entails extensive GPU-hours, substantial memory requirements, and a non-trivial infrastructure footprint. Even when deployed solely for inference, these models carry embedded carbon debt from their pretraining phase. A pragmatic approximation might allocate a fractional share (e.g., 1/1000\textsuperscript{th}) of the original training emissions per deployment use case, as suggested by prior studies~\cite{lacoste2019quantifying, henderson2020towards}. While such amortized lifecycle costs are not included in our current figures, they would further accentuate the relative efficiency of lightweight, task-specific models such as CNN3D in repeated or distributed settings.

In addition, centralized training paradigms—particularly with high-resolution, sensitive video data—introduce often-overlooked network energy costs. Uploading raw surveillance footage from edge devices to a central server for model training incurs significant bandwidth and power demands, especially when multiplied across thousands of geographically distributed clients. In contrast, federated learning transmits only compact model updates (e.g., LoRA adapter weights or personalized CNN parameters), dramatically reducing transfer overhead. The net benefit depends on network topology, data compression, and training frequency, but early estimates suggest that federated approaches can offer both privacy protection and lower communication-related emissions.

These considerations highlight that environmental sustainability in AI extends beyond algorithmic efficiency—it requires system-level thinking that includes model reuse, data logistics, and the full lifecycle impact of deployment architectures.

\section{Discussion}

Our comparison between LoRA-tuned LLaVA and a personalized CNN3D model highlights a critical trade-off for sustainable and privacy-aware AI deployment. Despite a nearly two orders of magnitude difference in parameter count—7 billion for LLaVA versus 65.8M for CNN3D—both models achieve comparable performance, surpassing 90\% accuracy and yielding similar F1 scores. Notably, CNN3D slightly outperforms in terms of ROC AUC and log loss, suggesting better calibration under non-IID conditions.

The environmental contrast is even more striking: CNN3D requires approximately \textbf{half the energy} (240 Wh vs. 570 Wh) and emits less than half the CO$_2$e (10.1 g vs. 24 g) compared to LoRA-based fine-tuning. These results underscore the effectiveness of parameter-frugal personalization for energy-conscious federated learning.

While LoRA-tuned vision-language models (VLMs) offer advantages in multimodal reasoning and interpretability, their higher computational cost makes them more suitable for selective use—e.g., in situations requiring nuanced scene understanding, long-range dependencies, or human-centric queries. In contrast, lightweight CNNs serve as reliable and efficient backbones for routine classification tasks in video surveillance.

These results motivate—but do not prescribe—a strategic hybrid deployment pattern, where lightweight CNNs handle routine inference by default, and transformer-based VLMs are selectively triggered in complex or ambiguous cases.

Future directions include developing context-aware model selection mechanisms, optimizing client-side inference latency, and extending evaluation to richer tasks such as video captioning, anomaly explanation, or real-time alert prioritization.

\section{Conclusion}

This study had two main objectives: (i) assess whether large vision-language models (VLMs) can operate efficiently in zero-shot or federated contexts, and (ii) evaluate whether lightweight, personalized models can offer comparable performance with lower environmental cost.

We conducted a systematic comparison between LoRA-tuned LLaVA-7B and a compact CNN3D model trained via personalized federated learning. Both were tested under realistic non-IID conditions, with explicit tracking of energy consumption and CO$_2$ emissions.

Key findings include:
\begin{itemize}
  \item Both models exceeded 90\% accuracy on validation sets;
  \item CNN3D slightly outperformed LLaVA in ROC AUC and log loss, while using less than half the energy;
  \item LoRA-tuned VLMs offered richer multimodal reasoning and interpretability;
  \item Zero-shot inference remains viable but energy and domain-sensitivity trade-offs must be considered.
\end{itemize}

These results highlight the value of hybrid deployment strategies: using efficient CNNs for routine inference, and selectively activating VLMs for complex or descriptive scenarios. They also reinforce the importance of frugal design beyond training—considering inference efficiency and lifecycle emissions.

Future work under the \textbf{DIVA} initiative will explore adaptive model selection, multimodal fusion (e.g., audio–text), and privacy-aware optimization on edge devices. We also aim to expand evaluation frameworks to include structured outputs and full lifecycle analysis.

% \section*{Bibliography}

\end{document}